\documentclass[letterpaper]{article}

\usepackage[preprint]{aaai2027}
\usepackage[hyphens]{url}
\usepackage{graphicx}
\usepackage{natbib}
\usepackage{caption}
\usepackage{amsmath}
\usepackage{amssymb}
\usepackage{algorithm}
\usepackage{algorithmic}
\usepackage{booktabs}

\title{Workflow-Localized Mechanism Learning: Attribution-Guided Repair and Knowledge Reuse for Structured Agent Skills}
\author{%
\begin{tabular}{@{}c@{\hspace{1em}}c@{\hspace{1em}}c@{}}
Zibin Lin & Shengli Zhang\corresponding & Taotao Wang\\[-0.05em]
{\small\normalfont linaacc9595@gmail.com} &
{\small\normalfont zsl@szu.edu.cn} &
{\small\normalfont ttwang@szu.edu.cn}\\[0.4em]
Yihan Xia & Deen Ma & Guofu Liao\\[-0.05em]
{\small\normalfont xiayihan2023@email.szu.edu.com} &
{\small\normalfont madeen2025@mails.szu.edu.cn} &
{\small\normalfont liaoguofu2022@email.szu.edu.cn}
\end{tabular}%
}
\affiliations{Shenzhen University, Shenzhen, China}

\begin{document}
\maketitle

\begin{abstract}
Agent Skills package reusable procedural knowledge as external artifacts for frozen language-model agents, yet existing optimizers do not jointly resolve where a failure occurs in a workflow, which mechanism caused it, and how relevant knowledge from third-party Skills should be reused locally. We introduce \emph{Workflow-Localized Mechanism Learning} (WML). Its Node--Mechanism Attribution identifies the failed workflow node, implicated mechanisms, and smallest valid edit target, routing single-mechanism defects to L3 resources and relational defects across mechanisms to L2 composition protocols. A six-module \emph{Workflow-Guided Skill Optimization} (WGSO) loop then selects provenance- and scope-aware third-party knowledge, applies bounded patches, evaluates candidates, and stores verified outcomes in optimizer-side memory. On SpreadsheetBench, WML reaches $90.33{\pm}1.53$ and $74.67{\pm}3.51$ Hard Accuracy with DeepSeek and Qwen3.6-Flash, respectively; without additional optimization, the learned Skills transfer to WikiTableQuestions with $84.00{\pm}2.00$ and $83.00{\pm}2.00$ Denotation Accuracy. On Compiler-Supported50, WML attains both the highest hard-PASS rate and the lowest cost per successful task; compiled execution sharply reduces tokens and calls relative to a direct SkillAgent while retaining most of its successful tasks. Code and artifacts are available at \url{https://github.com/xiaolin9595/workflow-localized-mechanism-learning}.
\end{abstract}

\section{Introduction}

Language-model agents can read files, call tools, execute code, and revise actions using verifier feedback \citep{yao2023react,schick2023toolformer,wang2024voyager,yang2024sweagent}. Their task competence, however, depends not only on model weights or a one-off prompt but also on reusable procedures, tool policies, failure modes, and verification rules. Agent Skills package this procedural knowledge as portable artifacts and use progressive disclosure to separate an always-loaded workflow from conditionally loaded resources \citep{li2026skillsbench,jiang2026sok,agentskills2026spec}.

Existing third-party Skills provide a second, complementary learning signal: procedures, decision rules, fallbacks, and verification patterns encoded by other developers or prior tasks. Reusing them can prevent an optimizer from rediscovering guidance from a small set of failures. Yet third-party Skills differ in assumptions, file organization, triggers, scope, and tool environments. The key problem is therefore not retrieving more text, but determining what knowledge the current failure lacks, which candidate is applicable, and where it should be adapted within the target Skill. Third-party reuse is itself a localized repair problem.

SkillOpt treats an external Skill document as trainable state for a frozen agent and performs bounded text edits from trajectory feedback \citep{yang2026skillopt}. SkillGrad extends this view to a multi-file package with textual diagnosis, cross-round aggregation, and layer-aware patches \citep{wang2026skillgrad}. Layer awareness answers \emph{which package layer} to edit, but leaves ambiguity inside a workflow: one stage may involve input semantics, target constraints, operation selection, commit contracts, and postconditions. A generic patcher must simultaneously infer the missing knowledge and its local protocol, which can turn a narrow failure into a global rewrite, duplicate an existing resource, or place task-specific rules in the always-loaded workflow.

Consider two failures at the same state-commit node. If an agent has acquired context, bound the target, and completed a transformation but commits the result in a representation forbidden by the task contract, a single output-contract mechanism lacks guidance; the smallest target is its L3 resource. If serialization, commit ordering, recovery, and post-commit verification rules already exist but their coordination is incomplete, expanding any one resource cannot repair the relation; the smallest target is an L2 composition protocol. The failures share a node but require different attribution states and update contracts.

We propose \emph{Workflow-Localized Mechanism Learning} (WML), which refines structured Skill optimization into workflow-localized diagnosis and knowledge reuse. WML represents the failed node, reusable mechanisms, defect relation, and layer-aware edit target through \emph{Node--Mechanism Attribution}. Its six-module implementation, \emph{Workflow-Guided Skill Optimization} (WGSO), converts execution evidence into a local knowledge need; consults third-party Skills only when the deployed Skill and optimizer memory are insufficient; and admits a bounded patch only after post-patch evaluation. External records never enter the deployed artifact verbatim, while a Patch-Strategy Memory (PSM) distills verified edit outcomes for future patching without exposing optimizer state to the executor.

Our contributions are threefold:
\begin{itemize}
    \item We formulate Node--Mechanism Attribution, moving Skill repair from document- or layer-level editing to typed localization over workflow nodes, mechanism relations, and L2/L3 edit addresses.
    \item We introduce WGSO, a closed loop that combines attribution-conditioned reuse of provenance-scoped third-party knowledge, bounded patching, post-patch evaluation, and optimizer-side strategy memory.
    \item We establish consistent gains across two backbones and two benchmarks, component-wise degradation across all ablation seeds, and favorable success--efficiency trade-offs under a unified LLM workflow compiler.
\end{itemize}

The remainder of the paper is organized as follows. The Related Work section positions WML within textual Skill optimization, workflow systems, third-party knowledge reuse, and cross-round repair. The Problem Formulation section defines structured Skills, execution evidence, and Node--Mechanism Attribution. The Workflow-Localized Mechanism Learning section presents WGSO's localized repair paths, knowledge-reuse procedure, optimizer memory, and prototype implementation. The Experimental Setup section describes the benchmarks, baselines, implementation details, and evaluation protocol, while the Results section reports the main, transfer, ablation, and efficiency findings. Finally, the Limitations and Ethical Considerations section discusses the scope of the evidence and responsible deployment, and the Conclusion summarizes the findings and implications.

\section{Related Work}

\subsection{Text and Structured Skill Optimization}

TextGrad and GEPA use language feedback as an update signal for prompts or textual programs \citep{yuksekgonul2025textgrad,agrawal2026gepa}. SkillOpt applies bounded edits, validation, and optimizer-side memory to external Skill documents \citep{yang2026skillopt}; SkillGrad optimizes progressively disclosed Skill packages through textual diagnosis, cross-round aggregation, and layer-aware patching \citep{wang2026skillgrad}. Trace2Skill distills trajectory-local lessons, while EvoSkill evolves candidate Skill artifacts \citep{ni2026trace2skill,alzubi2026evoskill}. WML does not re-claim that Skills are trainable state or that edits should be layer-aware. It addresses the unresolved target ambiguity inside a layer: which workflow node failed, whether the defect lies in a mechanism or its relation to others, and what the smallest valid address is.

\subsection{Workflows and Third-Party Knowledge Reuse}

Voyager, DSPy, AFlow, and WorkflowLLM organize agent behavior through skill libraries, program compilation, or workflow search \citep{wang2024voyager,khattab2024dspy,zhang2025aflow,fan2025workflowllm}. SkillFoundry and SkillX build reusable Skill knowledge bases from heterogeneous resources \citep{shen2026skillfoundry,wang2026skillx}. These systems demonstrate the value of external and historical knowledge, but library construction alone does not specify what a current structured-Skill failure lacks, whether a candidate applies, or which package address should change. WGSO treats retrieval as evidence for a localized repair rather than concatenating an entire external Skill into context.

\subsection{Cross-Round Diagnosis and Repair Strategy}

SkillGrad's persistent pattern memory aggregates recurring diagnosis directions and tracks whether the deployed Skill has absorbed corresponding guidance \citep{wang2026skillgrad}. WML retains diagnostic memory but separates it from a second optimizer-side state: the edit kind, scope, preservation constraints, and verification requirements that succeeded for a localized defect. Node--Mechanism Attribution identifies the repair target, WGSO executes it, and PSM learns an attribution-conditioned repair strategy from evaluated outcomes.

\section{Problem Formulation}

\subsection{Structured Skills and Execution Evidence}

Following SkillGrad, the structured Skill exposed to the executor at round $t$ is
\begin{equation}
S_t=(S_t^{L1},S_t^{L2},S_t^{L3}),
\end{equation}
where $S_t^{L1}$ contains discovery and activation metadata, $S_t^{L2}$ is the always-loaded workflow body, and $S_t^{L3}=\{r_1,\ldots,r_m\}$ contains mechanism resources loaded on demand. The optimizer separately maintains memory $M_t$ and a procedural-knowledge index $K_t$, neither of which is distributed with the deployed artifact. $K_t$ is initialized from an external Skill corpus $\mathcal C$ and stores retrievable records with provenance and applicability boundaries.

Given a frozen executor $F$, the current Skill, and training batch $B_t$, execution and assessment produce
\begin{equation}
E_t=\operatorname{ExecuteAssess}(F,S_t,B_t),
\end{equation}
including outcomes, trajectories, evaluator feedback, and successful evidence used as preservation constraints. WML abstracts tool-agent execution into six domain-general functional nodes: context acquisition, target binding, operation selection, execution/transformation, state commit, and verification/termination. These describe roles rather than spreadsheet-specific objects; each domain instantiates them with its own state and tools.

\subsection{Node--Mechanism Attribution}

For task-level diagnosis $D_{t,i}$, WML defines the attribution state
\begin{equation}
A_{t,i}=(n_{t,i},\Gamma_{t,i},\rho_{t,i},\tau_{t,i}),
\end{equation}
where $n_{t,i}\in\mathcal N$ is the failed workflow node, $\emptyset\neq\Gamma_{t,i}\subseteq\mathcal V_{\mathrm{mech}}$ is the diagnosed reusable-mechanism set, $\rho_{t,i}\in\{\text{single-guidance},\text{multi-relation}\}$ is the defect relation, and $\tau_{t,i}$ is a typed edit address containing a layer, file, and section. The WML schema imposes the routing contract
\begin{align}
\rho_{t,i}=\text{single-guidance}
&\Rightarrow |\Gamma_{t,i}|=1,\quad
\tau_{t,i}\in\operatorname{Addr}(S_t^{L3}), \nonumber\\
\rho_{t,i}=\text{multi-relation}
&\Rightarrow |\Gamma_{t,i}|\geq2,\quad
\tau_{t,i}\in\operatorname{Addr}(S_t^{L2}).
\end{align}
An L3 address points to the resource implementing one mechanism. An L2 address points to a composition protocol coordinating resource selection, ordering, scope or conflict handling, and joint verification at the same node. This correspondence is a typed contract of our structured-Skill schema, not a claim that mechanism count universally determines a package layer. Implementations also attach supporting evidence and edit-scope constraints as audit fields.

\subsection{Workflow-Localized Update}

Compatible attribution records are aggregated into local repair proposals. For proposal $a=(D_a,n_a,\Gamma_a,\rho_a,\tau_a)$, memory context $C_t$, and selected knowledge $X_t$, write SG for single-guidance and MR for multi-relation. The patcher proposes
\begin{equation}
\widehat S_{t+1}=\begin{cases}
\operatorname{Patch}_{L3}(S_t,a,X_t,C_t), & \rho_a=\mathrm{SG},\\
\operatorname{Patch}_{L2\text{-comp}}(S_t,a,X_t,C_t), & \rho_a=\mathrm{MR}.
\end{cases}
\end{equation}
Only the attributed address may change, and reusable behavior from successful trajectories becomes a preservation constraint. The candidate is re-evaluated on the same batch, $\widehat E_t=\operatorname{ExecuteAssess}(F,\widehat S_{t+1},B_t)$, and an evaluation gate accepts, rejects, or rolls back the update:
\begin{equation}
(g_t,S_{t+1})=\operatorname{EvaluateGate}(S_t,\widehat S_{t+1},E_t,\widehat E_t).
\end{equation}
This is structured textual repair constrained by execution evidence, rather than differentiable credit assignment.

\section{Workflow-Localized Mechanism Learning}

\subsection{WGSO Overview}

\begin{figure*}[t]
\centering
\includegraphics[width=\textwidth]{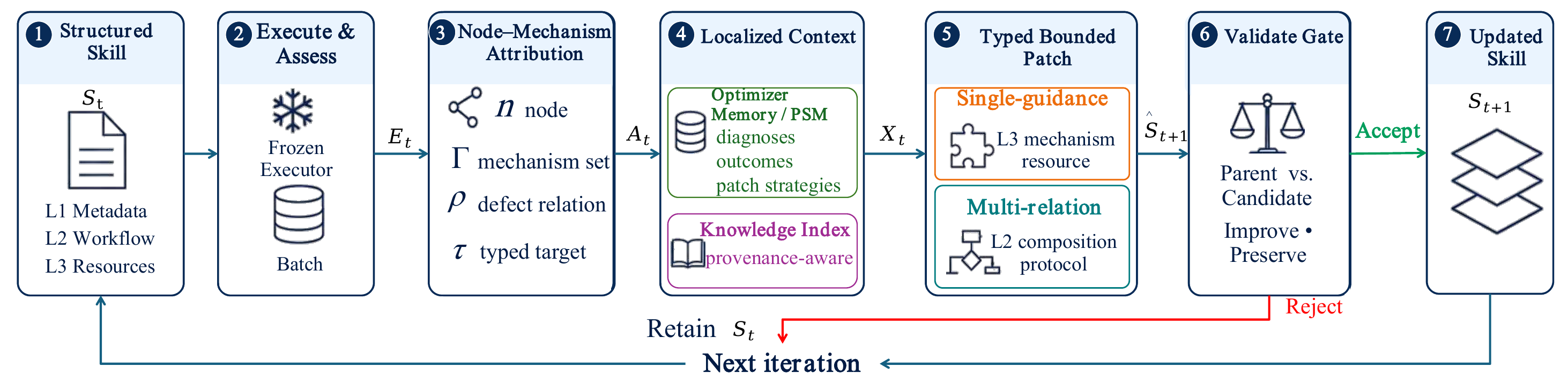}
\caption{WGSO overview. Node--Mechanism Attribution maps evidence to $(n,\Gamma,\rho,\tau)$, routing single-guidance defects to an L3 resource and multi-relation defects to an L2 composition protocol. Only evaluated improvements update the deployed Skill; external records and PSM remain optimizer-side.}
\label{fig:wgso}
\end{figure*}

WGSO realizes WML through the transition
\begin{equation}
(S_{t+1},M_{t+1},K_{t+1})=\operatorname{WGSO}(S_t,M_t,K_t,B_t;F).
\end{equation}
As Figure~\ref{fig:wgso} shows, execution evidence is attributed to a typed local repair; optimizer memory and third-party records supply bounded context; the candidate is validated and evaluated; and the outcome updates both the deployed Skill and optimizer-side state.

\subsection{Attribution and Dual-Path Repair}

Node--Mechanism Attribution first identifies the functional node, then the reusable mechanisms and their defect relation. For a state-commit failure caused by an invalid result representation, it yields state commit $\rightarrow$ output contract $\rightarrow$ single guidance and targets the corresponding L3 resource. The patch tightens only that mechanism's decision rule, scope boundary, and verification requirement. If serialization, ordering, recovery, and post-commit verification already exist but are composed incorrectly, attribution targets their multi-relation defect and patches the L2 composition protocol without copying or merging the L3 procedures. Both paths enforce a minimum edit scope; the optimizer performs no update when guidance already covers the diagnosis, attributions conflict, or evidence is insufficient.

Operationally, the diagnoser receives the task outcome, relevant trajectory spans, grader feedback, the current L2/L3 contents, and successful evidence that must be preserved. It proposes mechanism anchors together with file/section evidence. A deterministic registry then maps each anchor to legal workflow-node and package-address candidates, normalizing free-form diagnoses and preventing the model from selecting an arbitrary file. The guarded resolver classifies the defect by testing guidance coverage: a missing or incorrect rule internal to one mechanism is \emph{single-guidance}, whereas individually available mechanisms with missing selection, ordering, scope/conflict handling, or joint verification form a \emph{multi-relation} defect. It assigns $\tau$ to the smallest registered section capable of expressing the missing rule.

Each attribution retains the supporting trajectory or evaluator span and preservation constraints used to justify the decision. When the diagnoser proposes multiple locations, the resolver intersects them with registry-valid addresses and selects the most local compatible target. It abstains when that intersection is empty, candidate records imply inconsistent nodes or relations, existing guidance already covers the evidence, or the evidence does not support a unique legal edit scope. Thus $A_{t,i}$ is both a repair decision and an audit record linking observed failure evidence to the permitted patch surface.

\subsection{Localized Reuse of Third-Party Skills}

WGSO treats third-party Skills as sources of procedural knowledge, not documents to append to the prompt. During initialization, it converts reusable content from $\mathcal C$ into records
\begin{equation}
x=(u,m,s,p,v,\gamma),
\end{equation}
where $u$ is provenance, $m$ is a mechanism, $s$ is the applicability scope, $p$ is a procedure or decision rule, $v$ is a verification requirement, and $\gamma$ is confidence. Full records remain in the optimizer-side index $K_t$.

For a diagnosis, WGSO first checks Skill coverage and memory. Only when guidance for the attributed mechanism is insufficient does retrieval select a bounded $X_t\subseteq K_t$ using the mechanism or anchor, pattern phrase, scope, and past edit outcomes. Corpus-backed records retain source and scope. If no record matches, the system may form a diagnosis-derived provisional candidate, but does not treat it as external evidence. Selected records are adapted, merged, or rejected by the patcher under the attribution target and preservation constraints; they affect $S_{t+1}$ only after the gate accepts the candidate. Thus retrieval decides \emph{what} knowledge may help, while typed routing controls \emph{where} it may be written.

\subsection{Optimizer Memory and PSM}

$M_t$ stores recurring diagnoses, edit outcomes, and PSM. Diagnosis records track repeated mechanisms and whether the Skill has absorbed the corresponding guidance. Outcome records track acceptance, rejection, and rollback. PSM instead summarizes which edit kind, scope, preservation constraint, and verification requirement succeeded for a localized defect. It updates only after verified improvement and remains separate from the executor-facing Skill. External records answer what guidance can be reused; PSM answers how to apply it safely under the current attribution.

A PSM entry is indexed by a normalized attribution signature consisting of the workflow node, mechanism set, defect relation, and target layer. Its value records the successful edit kind, permitted section scope, preservation constraints, required postconditions, and evaluated outcome. Before patching, \textsc{ReadMemory} retrieves entries with compatible signatures and exposes them only to the patcher, where they constrain the edit plan and verification checklist. A verified improvement inserts or reinforces the corresponding strategy. Rejected or regressed candidates remain available as negative edit-outcome records in $M_t$ but do not update PSM. Consequently, PSM can discourage a previously harmful edit without placing optimizer history or third-party text in the deployed Skill.

\subsection{Algorithm and Prototype Instantiation}

Algorithm~\ref{alg:wgso} formalizes WGSO. Each round converts execution evidence into workflow-localized proposals, augments only guidance-deficient proposals, routes them through the typed L3/L2 repair paths, and commits a candidate only after validation and evaluation.

\begin{algorithm}[t]
\caption{Workflow-Guided Skill Optimization (WGSO)}
\label{alg:wgso}
\begin{algorithmic}[1]
\STATE \textbf{Input:} initial Skill $S_0$; frozen executor $F$; training set $\mathcal D$; external corpus $\mathcal C$; rounds $T$
\STATE \textbf{Output:} selected Skill $S_T$ and optimizer state $(M_T,K_T)$
\STATE Validate $S_0$; $M_0\leftarrow\textsc{InitMemory}()$
\STATE $K_0\leftarrow\textsc{BootstrapIndex}(\mathcal C)$
\FOR{$t=0,\ldots,T-1$}
  \STATE $B_t\leftarrow\textsc{SampleBatch}(\mathcal D,t)$
  \STATE $E_t\leftarrow\textsc{ExecuteAssess}(F,S_t,B_t)$
  \STATE $\mathcal A_t\leftarrow\textsc{NodeMechanismAttribution}(E_t,S_t)$
  \STATE $\mathcal P_t\leftarrow\textsc{Aggregate}(\mathcal A_t)$
  \STATE $C_t\leftarrow\textsc{ReadMemory}(M_t,\mathcal P_t)$
  \IF{$\textsc{GuidanceInsufficient}(S_t,\mathcal P_t,C_t)$}
    \STATE $(X_t,K_{t+1})\leftarrow\textsc{RetrieveExtract}(K_t,\mathcal P_t)$
  \ELSE
    \STATE $(X_t,K_{t+1})\leftarrow(\varnothing,K_t)$
  \ENDIF
  \STATE $\widehat S_{t+1}\leftarrow S_t$
  \FOR{each $a\in\mathcal P_t$}
    \IF{$\rho_a=\mathrm{SG}$}
      \STATE $\widehat S_{t+1}\leftarrow\operatorname{Patch}_{L3}(\widehat S_{t+1},a,X_t,C_t)$
    \ELSE
      \STATE $\widehat S_{t+1}\leftarrow\operatorname{Patch}_{L2\text{-comp}}(\widehat S_{t+1},a,X_t,C_t)$
    \ENDIF
  \ENDFOR
  \STATE $\Delta_t\leftarrow\textsc{StructuredDiff}(S_t,\widehat S_{t+1})$
  \STATE $V_t\leftarrow(F,S_t,\widehat S_{t+1},B_t,E_t,\Delta_t)$
  \STATE $(g_t,S_{t+1},\widehat E_t)\leftarrow\textsc{ValidateAndGate}(V_t)$
  \STATE $H_t\leftarrow(\mathcal A_t,\mathcal P_t,X_t,\Delta_t,g_t,\widehat E_t)$
  \STATE $M_{t+1}\leftarrow\textsc{UpdateMemory}(M_t,H_t)$
\ENDFOR
\STATE \textbf{return} $S_T,M_T,K_T$
\end{algorithmic}
\end{algorithm}

Our prototype implements attribution as the auditable hybrid decision chain described above. The resulting $A_{t,i}$ controls both patch routing and locality. An L3 patch may modify only the attributed mechanism resource and its named section; it may refine a decision rule, scope boundary, or verification requirement but cannot alter another workflow node. An L2 patch may modify only the attributed composition protocol; it can coordinate L3 resources but cannot rewrite or duplicate their procedures. A structured diff checks the candidate against $\tau$ and the preservation constraints, making the edit boundary mechanically inspectable.

\textsc{ValidateAndGate} first rejects an incomplete candidate or a diff outside the permitted address. It then executes the candidate on the same batch and compares task- and cell-level outcomes with $E_t$ under the configured score guard. A structurally valid candidate is committed only after post-patch evaluation verifies improvement; an incomplete, invalid, or regressed candidate restores $S_t$. The gate records accepted, rejected, or regressed outcomes in general optimizer memory, while only verified improvements update PSM. The non-evolutionary path generates one candidate per round. The no-attribution ablation removes this entire attribution-constrained repair path.

\section{Experimental Setup}

\subsection{Research Questions and Datasets}

We ask: \textbf{RQ1}, does WML improve end-to-end performance across backbones and transfer without further optimization? \textbf{RQ2}, what do Node--Mechanism Attribution, procedural-knowledge enhancement, and PSM contribute? \textbf{RQ3}, how do final Skills compare in success rate and cost under a unified compiler, and how does constrained execution trade accuracy for efficiency against an open-ended SkillAgent?

SpreadsheetBench requires an agent to read an input workbook, produce an output workbook, and preserve non-target content \citep{ma2024spreadsheetbench}. Hard Accuracy requires every required cell in a task to be correct; Cell Accuracy averages correctness over annotated cells. For WikiTableQuestions random-split-1-dev \citep{pasupat2015wikitq}, we write the table into a workbook and ask the agent to place the denotation in a fixed answer cell. This is a cross-benchmark, zero-additional-training transfer test within spreadsheet-form execution, not a claim of general-domain out-of-distribution transfer.

\subsection{Baselines and Optimization Protocol}

We compare No Skill, a common Seed Skill, Official SkillGrad, SkillOpt, EvoSkill, and WML. Training-based methods receive the same number of optimization examples and are evaluated with the same frozen set, executor backbone, and evaluator; their batch construction, selection, and trajectory reuse follow their native algorithms. This controls data scale and evaluation while preserving the optimization strategy being compared. We use DeepSeek-chat with thinking disabled and Qwen3.6-Flash. Training-based results are means and sample standard deviations over three independent runs.

DeepSeek WML uses 40 optimization examples, batch size 4, five rounds, and a fixed 100-task evaluation. Executor, diagnoser, patcher, and memory updater use the same backbone within a run. Five rounds were fixed for the main study and all ablations. A separate ten-round diagnostic run tests longer optimization: round 5 first reaches 0.90 Hard and 0.9510 Cell Accuracy; doubling the budget leaves Hard at 0.90 and raises Cell Accuracy only to 0.9526. Round 5 is therefore the predefined operating point balancing performance, cost, and stability, not a post-hoc checkpoint selected from final evaluation (Figure~\ref{fig:budget}).

All ablations use DeepSeek-chat, no thinking, seeds 42/123/456, 40 optimization examples, batch size 4, five rounds, and 100 evaluation tasks per seed. We report mean $\pm$ sample SD using denominator $n-1$. Because there are three seeds, we report descriptive dispersion and mean differences rather than significance tests. Main and ablation results come from independent run sets. Cell values are rounded to four decimals before variance calculation.

The primary cross-backbone and cross-benchmark comparison is reported in Table~\ref{tab:main}.

\begin{table*}[t]
\centering
\small
\begin{tabular}{lcccc}
\toprule
& \multicolumn{2}{c}{DeepSeek} & \multicolumn{2}{c}{Qwen3.6-Flash} \\
\cmidrule(lr){2-3}\cmidrule(lr){4-5}
Method & SpreadsheetBench & WikiTableQuestions & SpreadsheetBench & WikiTableQuestions \\
\midrule
No Skill & 83.00 & 77.00 & 63.00 & 24.00 \\
Seed Skill & 82.00 & 78.00 & 70.00 & 77.50 \\
Official SkillGrad & $82.00{\pm}1.00$ & $81.00{\pm}1.00$ & $66.00{\pm}3.61$ & $76.00{\pm}3.00$ \\
SkillOpt & $81.67{\pm}1.53$ & $81.67{\pm}2.52$ & $60.00{\pm}3.61$ & $80.00{\pm}3.00$ \\
EvoSkill & $84.33{\pm}2.08$ & $81.67{\pm}2.52$ & $55.33{\pm}3.51$ & $76.67{\pm}2.52$ \\
\textbf{WML (Ours)} & $\mathbf{90.33{\pm}1.53}$ & $\mathbf{84.00{\pm}2.00}$ & $\mathbf{74.67{\pm}3.51}$ & $\mathbf{83.00{\pm}2.00}$ \\
\bottomrule
\end{tabular}
\caption{Primary results across backbones and benchmarks (\%). SpreadsheetBench reports Hard Accuracy; WikiTableQuestions reports zero-additional-training Denotation Accuracy. Training-based methods report mean $\pm$ sample SD over three runs.}
\label{tab:main}
\end{table*}

\subsection{Unified Compiler Protocol}

All final Skills are passed to one LLM workflow compiler. Given the same task prompt and environment summary, the compiler directly reads the input Skill and uses a shared model, prompt, and decoding configuration to generate a typed workflow specification and its executable workflow code. The specification exposes selected procedures, constraints, dependencies, and postconditions; code generation then realizes that specification against the common runtime interface. Every output passes the same schema and static checks, runs in the same sandboxed runtime, and is scored by the same grader. No method receives a custom prompt, mapping, adapter, code template, historical plan, answer injection, or WML-specific runtime rule. The only changing input is the final Skill artifact, so the protocol measures differences induced by Skill content, organization, and their interaction with one automated generation-and-execution stack.

Compiler-Supported50 is a fixed capability slice constructed once from WML development trajectories for which the compiler produced runnable artifacts. The same 50 task identifiers, prompts, workbook inputs, and evaluation manifest are then held fixed for every Skill; compiler, planner, runtime, grader, call budget, and retry policy also remain unchanged. This creates a controlled comparison inside the compiler-supported operating envelope: each method must expose useful procedural knowledge through its final Skill, while no method-specific adapter can compensate for missing or poorly organized guidance.

Success-normalized cost charges every attempted task, including failures. For a Skill $s$, tokens per hard PASS equal the total recorded planner tokens over all 50 tasks divided by its number of hard-PASS tasks; calls per hard PASS use the analogous total number of planning calls. A direct-execution control gives the same WML Skill to the native SkillGrad SkillAgent, which retains open-ended planning and arbitrary code generation. The compiled row counts planning calls and the direct row counts model requests, both interpreted as stack-level LLM invocations; same-Skill rows report their unnormalized totals.

\section{Results}

Table~\ref{tab:main} presents the primary end-to-end comparison before auxiliary analyses.

\subsection{End-to-End Performance and Transfer}

WML ranks first in all four columns of Table~\ref{tab:main}. Against the strongest non-WML method in each column, it improves DeepSeek SpreadsheetBench, DeepSeek WikiTableQuestions, Qwen3.6-Flash SpreadsheetBench, and Qwen3.6-Flash WikiTableQuestions by 6.00, 2.33, 4.67, and 3.00 percentage points, respectively. Against Official SkillGrad, the closest structured-Skill optimizer, the margins are 8.33, 3.00, 8.67, and 7.00 points. Gains hold for two executors, in-domain spreadsheet manipulation, and transfer to a separately constructed table-QA benchmark, indicating that the optimized procedural state is not tied to a single backbone or benchmark instance.

\begin{figure}[t]
\centering
\includegraphics[width=\columnwidth]{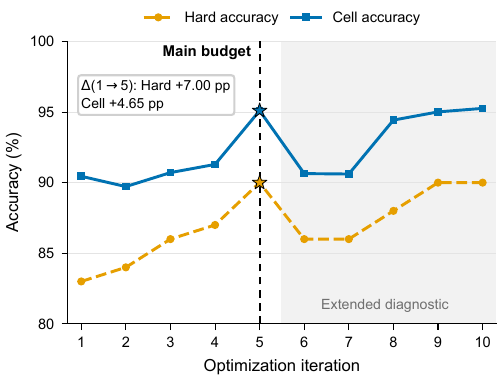}
\caption{Checkpoint dynamics under the predefined five-round budget and a separate extended diagnostic. The dashed line marks the main budget endpoint; rounds 6--10 are not used to select the reported main result.}
\label{fig:budget}
\end{figure}

\begin{figure}[t]
\centering
\includegraphics[width=\columnwidth]{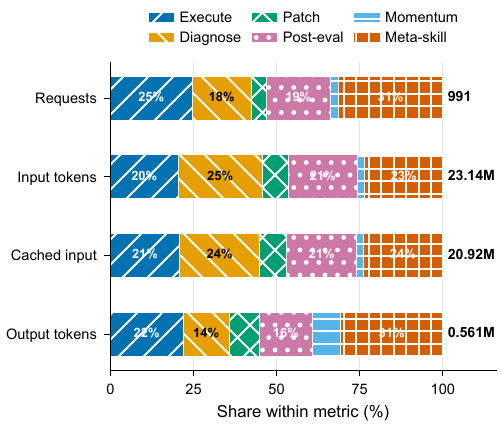}
\caption{Stage-level composition of WML's offline optimization cost. Each bar normalizes one cost category across six stages; cached input is a subset of input tokens.}
\label{fig:cost}
\end{figure}

Figure~\ref{fig:budget} supports the five-round choice: the additional five rounds do not produce a sustained gain commensurate with twice the optimization budget. One complete WML optimization uses 991 requests, 23.14M input tokens, and 0.561M output tokens; 20.92M input tokens are cached (90.42\%). Figure~\ref{fig:cost} shows that this is an offline artifact-construction cost. The resulting Skill is reusable by downstream executors without repeating optimization.

\subsection{Component Ablations}

Table~\ref{tab:ablation} evaluates three independent run sets under an otherwise identical protocol. Full WML outperforms every ablation at every seed. Removing Node--Mechanism Attribution lowers mean Hard/Cell Accuracy by 5.33/7.89 points, the largest Cell degradation, and introduces $2{\pm}1$ retries. This supports attribution-constrained locality as a central mechanism for precise repair. Disabling the full procedural-knowledge enhancement path costs 5.33/5.53 points, showing that failure-conditioned knowledge supplementation is necessary rather than incidental. Removing the patch-strategy meta-policy still costs 3.33/1.01 points, demonstrating an additional benefit from learning how evaluated local edits should be applied across rounds.

\begin{center}
\centering
\small
\begin{tabular}{@{}lcc@{}}
\toprule
Variant & \shortstack{Hard / Cell $\uparrow$\\(mean $\pm$ SD)} & \shortstack{$\Delta$ Hard / $\Delta$ Cell\\(pp)} \\
\midrule
Full WML & \shortstack{$90.00{\pm}1.00$\\$95.78{\pm}0.01$} & \shortstack{---\\---} \\
w/o Attribution & \shortstack{$84.67{\pm}0.58$\\$87.89{\pm}0.01$} & \shortstack{$-5.33$\\$-7.89$} \\
w/o Knowledge Enh. & \shortstack{$84.67{\pm}0.58$\\$90.25{\pm}0.01$} & \shortstack{$-5.33$\\$-5.53$} \\
w/o Meta-Policy & \shortstack{$86.67{\pm}0.58$\\$94.77{\pm}0.01$} & \shortstack{$-3.33$\\$-1.01$} \\
\bottomrule
\end{tabular}
\captionof{table}{WML component ablations on SpreadsheetBench (\%) over seeds 42/123/456. ``Knowledge Enh.'' denotes Procedural-Knowledge Enhancement; ``Meta-Policy'' denotes the Patch-Strategy Meta-Policy.}
\label{tab:ablation}
\end{center}

\subsection{Compilable Skills: Cross-Skill Efficiency and Deployment Trade-Off}

Prior work has established compilation and automatic workflow generation as viable interfaces for operationalizing LM programs \citep{khattab2024dspy,zhang2025aflow,fan2025workflowllm}. We use this paradigm to test a broader systems hypothesis: can an optimized Skill serve as a procedural intermediate representation between knowledge learning and agent execution? The compiler is not part of WML optimization; it acts as a shared downstream consumer that must translate independently produced Skill artifacts into task-conditioned workflows and code.

Under the unified compiler, WML solves 40/50 tasks, 17 more than the strongest non-WML result---an absolute lead of 34 points---and leads Cell Accuracy by 21.30 points. It also uses 32.73\% fewer tokens per hard PASS and 48.28\% fewer calls per hard PASS than the corresponding strongest non-WML costs. Because every row shares the compiler and execution stack and only the input Skill changes, these results compare the operational quality of complete Skill artifacts rather than method-specific adapters.

The same-Skill control isolates execution mode. The direct SkillAgent completes six additional tasks and improves Cell Accuracy by 6.84 points. Compiled execution retains 40 of its 46 successful tasks (87.0\%) while reducing tokens by 84.90\% and stack-level LLM invocations by 90.39\%. The compiler therefore converts workflow knowledge into a lower-cost, constrained, and auditable execution protocol; the direct agent preserves stronger on-the-fly planning and arbitrary code generation at substantially higher cost.

\begin{center}
\centering
\small
\begin{tabular}{@{}lcc@{}}
\toprule
Skill or mode & \shortstack{Hard PASS\\Cell $\uparrow$} & \shortstack{Tokens $\downarrow$\\LLM calls $\downarrow$} \\
\midrule
\multicolumn{3}{l}{\emph{Cross-Skill: cost per hard PASS}} \\
\textbf{WML} & \shortstack{\textbf{40/50}\\\textbf{0.880695}} & \shortstack{\textbf{33,352/PASS}\\\textbf{1.35/PASS}} \\
EvoSkill & \shortstack{23/50\\0.667680} & \shortstack{49,576/PASS\\2.91/PASS} \\
SkillOpt & \shortstack{23/50\\0.631380} & \shortstack{54,360/PASS\\2.61/PASS} \\
Seed Skill & \shortstack{22/50\\0.629755} & \shortstack{51,135/PASS\\2.91/PASS} \\
Official SkillGrad & \shortstack{19/50\\0.637745} & \shortstack{60,342/PASS\\3.32/PASS} \\
\midrule
\multicolumn{3}{l}{\emph{Same WML Skill: total execution cost}} \\
Compiled executor & \shortstack{40/50\\0.8807} & \shortstack{1,334,063\\54} \\
Direct SkillAgent & \shortstack{46/50\\0.9491} & \shortstack{8,835,158\\562} \\
\bottomrule
\end{tabular}
\captionof{table}{Compiler-Supported50 cross-Skill efficiency and the execution-mode trade-off for the same WML Skill. Cross-Skill rows report cost per hard PASS; same-Skill rows report total cost.}
\label{tab:compiler}
\end{center}

Beyond the comparison of execution modes, these results motivate a division of labor for future agent systems. An optimizer can learn and organize reusable procedures once; a compiler can lower the resulting artifact into typed workflows and executable code; and a restricted runtime can enforce explicit checks and tool boundaries. This separation amortizes optimization cost across tasks, decouples procedural knowledge from a particular executor, and exposes plans, constraints, and privileges for auditing. Open-ended agents remain essential outside the compiler-supported envelope. The prospective architecture is therefore hybrid: covered tasks use efficient compiled execution, while novel cases fall back to direct agents and contribute new evidence for subsequent Skill improvement. Table~\ref{tab:compiler} provides an initial empirical basis for this paradigm: WML produces the most compiler-consumable Skill among the compared methods, while the same-Skill control quantifies the accuracy--efficiency frontier rather than assuming that either execution mode universally dominates.

\section{Limitations and Ethical Considerations}

Our empirical study focuses on spreadsheet agents. The six-node ontology, mechanism vocabulary, and L2/L3 schema require validation in software maintenance, browser control, and embodied tasks. WikiTableQuestions is converted to spreadsheet-form execution, so our evidence supports cross-benchmark transfer within this execution setting. Baselines receive equal optimization-data scale and a common evaluation protocol while retaining native training and selection strategies; the comparison evaluates complete optimizers rather than isolating each internal operation. Ablations use three seeds, and Compiler-Supported50 evaluates the compiler-supported capability domain. The current post-patch gate uses same-batch comparison; an independent long-horizon regression set is important future work.

The experiments use public benchmarks and procedural Skill artifacts and introduce no new human-subject data. Agents that execute code or modify workbooks can corrupt files, expose private data, or act on unauthorized content. Deployment should therefore sandbox execution, minimize file and network privileges, and log provenance, tool traces, and grader decisions. External procedural records should preserve source, license, scope, and contamination audits, and should be screened for prompt injection or malicious procedures before entering optimizer context.

\section{Conclusion}

We introduced WML, which converts a failed workflow node, mechanism relation, and L2/L3 target into an executable local repair decision. WGSO closes the loop through localized third-party knowledge reuse, bounded patching, outcome gating, and optimizer-side strategy memory. WML achieves the strongest mean result in all four backbone--benchmark combinations; removing attribution lowers Hard/Cell Accuracy by 5.33/7.89 points; and, on Compiler-Supported50, WML leads the strongest alternative by 17 hard-PASS tasks while delivering the best success-normalized efficiency. These results show that workflow-localized attribution and constrained knowledge reuse can produce structured procedural state that is both effective for agents and operationally consumable by a compiler, supporting a learn-once, compile-many path toward efficient and auditable agent systems.

\bibliography{wml-arxiv}

\end{document}